\documentclass[letterpaper]{article} 
\usepackage{aaai2026}  
\usepackage{times}  
\usepackage{helvet}  
\usepackage{courier}  
\usepackage[hyphens]{url}  
\usepackage{graphicx} 
\usepackage{natbib}  
\usepackage{caption} 
\usepackage{booktabs}
\usepackage{tabularx}
\usepackage{array}
\usepackage{amsmath}
\usepackage{comment}
\usepackage{pgfplots}
\pgfplotsset{compat=1.18} 

\definecolor{likert5}{HTML}{0D33D2}
\definecolor{likert4}{HTML}{4460D8}
\definecolor{likert3}{HTML}{7B8EDD}
\definecolor{likert2}{HTML}{B1BBE3}
\definecolor{likert1}{HTML}{E8E8E8}

\title{Enforcing Narrative Reliability and Epistemic Pacing in LLM-Driven Detective Games via Structured Knowledge Trees}
\author {
    Parsa Rahmati,
    Richard Zhao
}
\affiliations {
    Department of Computer Science, University of Calgary\\
    Calgary, Alberta, Canada, T2N 1N4\\
    parsa.rahmaty@ucalgary.ca, richard.zhao1@ucalgary.ca
}

\begin{document}

\maketitle

\begin{abstract}
Large Language Models (LLMs) enable open-ended dialogue in interactive games, but their non-deterministic outputs make it difficult to preserve authorial control, factual consistency, and the intended sequence of information disclosure. These challenges are particularly significant in detective games, where premature revelation or fabricated details can undermine the logic of player progression. We present a Structured Knowledge Tree architecture coupled with a tri-agent LLM pipeline for controlling dialogue in an open-ended interrogation game. The system separates knowledge retrieval, dialogue generation, and response verification to ensure that the virtual suspect reveals only information permitted by the current narrative state. We evaluate the approach through \textit{The Interrogation of Adrian Gale}, a playable detective-game testbed, and a formal user study examining hallucination reduction, adherence to authored disclosure sequences, and perceived logical progression. Our results demonstrate that the structured architecture reduces critical hallucinations by 64.78\% and entirely prevents premature narrative disclosure. While the strict mechanical constraints introduced usability trade-offs regarding forced conversational reveals, the system successfully enforces rigorous epistemic pacing and provides players with a clear, subjective sense of progression toward solving the case.
\end{abstract}

\section{Introduction}
The integration of Large Language Models (LLMs) has introduced unprecedented possibilities in video game design \cite{sweetser2024large}. These models bring us closer to the long-standing goal of creating interactive systems capable of open-ended, natural language communication with players. However, because LLM outputs are inherently non-deterministic, developers struggle to guarantee a consistent, authored experience \cite{sun2023language}. When utilized for narrative generation, LLMs frequently hallucinate unauthorized details that deviate from the narrative designer’s core vision \cite{treanor2024prototyping}. In genres that rely on rigid mechanics and sequential discovery, such as detective or mystery games, this instability breaks the logical order of information reveal, compromising the game state.

Recent mitigation strategies, such as symbolically grounding LLMs \cite{treanor2024prototyping} or optimizing Retrieval-Augmented Generation (RAG) pipelines \cite{chien2025reinforced, chen2025improving}, struggle to translate to the mystery genre because they are fundamentally designed to surface the complete truth as efficiently as possible. In a detective narrative, the objective of the virtual agent is radically different: the system must intentionally withhold information and reveal clues only in a logically meaningful order. Treating an LLM as an actor in a mystery requires epistemic pacing (the ability to deceive, deflect, and hide facts), which is a constraint that truth-maximizing or sandbox-oriented models are not equipped to enforce. Without strict control over this pacing, an agent might prematurely surrender critical clues, which destroys the investigative logic of the narrative and robs the player of a meaningful sense of progression.

To explore a potential solution to these challenges, we developed \textit{The Interrogation of Adrian Gale}, an interactive detective game centered on open-ended natural language interrogation. In this experience, players assume the role of a detective tasked with extracting critical clues and ultimately securing a confession by typing conversational queries to a virtual suspect. Rather than relying on a single generative model to track the narrative, we utilize a Structured Knowledge Tree (SKT) to store the game's story and prerequisite logic. While knowledge graphs are established methods for organizing structured information \cite{hogan2021knowledge}, our approach investigates whether coupling this static structure with a specialized, multi-agent LLM pipeline can effectively enforce narrative pacing in real-time. We implemented a tri-agent system to interface with the SKT, separating logic, performance, and verification: an Analysis LLM navigates the tree to find permissible statements, a Dialogue LLM generates the character's dialogue based strictly on those constraints, and a Detection LLM verifies whether the information was accurately conveyed.

We evaluated this approach through a formal player study comparing it to standard generative models, specifically addressing the following questions:

\begin{itemize}
    \item \textbf{RQ1:} To what extent does the SKT architecture reduce LLM hallucinations and ensure the generated dialogue remains accurate to the authored story?
    \item \textbf{RQ2:} How effectively does the tri-agent pipeline control narrative pacing and prevent premature confessions compared to an LLM-Only version?
    \item \textbf{RQ3:} How does the structured pacing of the SKT impact the player's perceived sense of logical discovery and game progression?
\end{itemize}
To address these questions, this paper presents two primary contributions:
\begin{itemize}
    \item \textbf{The Tri-Agent SKT Architecture:} The design and implementation of a novel LLM pipeline that uses a Structured Knowledge Tree to enforce narrative pacing and prevent premature information reveals.
    \item \textbf{Empirical Evaluation of Epistemic Pacing:} A formal player study using our custom playable testbed, \textit{The Interrogation of Adrian Gale}. The results demonstrate that our structured architecture effectively mitigates unauthored narrative deviations and prevents premature confessions, all while maintaining the player's subjective sense of logical discovery.
\end{itemize}

\section{Related Works}

\subsection{Interactive Drama and Narrative Control}
Riedl and Bulitko (\citeyear{riedl2013interactive}) explore the inherent tension between authorial intent and player agency, identifying a spectrum of architectural approaches ranging from strong-story to strong-autonomy. Strong-story systems preserve authorial intent through rigid branching paths, whereas strong-autonomy frameworks, like the usage of Hierarchical Task Networks (HTNs) in Cavazza et al. (\citeyear{cavazza2002interacting}), allow narratives to emerge organically by having characters dynamically re-plan around player interventions. Seeking a middle ground, hybrid architectures like the seminal \textit{Fa\c{c}ade} \cite{mateas2002behavior} successfully balanced agency and plot by sequencing discrete narrative ``beats'' in response to parsed player text. While \textit{Fa\c{c}ade} provided unprecedented conversational agency for its time, modern LLMs can now cover a vastly wider range of natural language input. However, this open-endedness introduces severe pacing risks for epistemic narratives (e.g., mysteries), which demand rigid rules around information disclosure \cite{ryan2008interactive}. Unlike dynamic re-planning or template-based parsing, our work explores how to safely harness LLM flexibility by enforcing strict epistemic boundaries, preventing the narrative from dynamically unspooling while still offering free-form conversational agency.

\subsection{Knowledge Representation and Boundary Enforcement}
To mitigate LLM hallucinations and enforce authorial control, researchers increasingly ground generative text in external data structures. The foundational approach, Retrieval-Augmented Generation (RAG) \cite{lewis2020retrieval}, retrieves semantically relevant passages from an external database and appends them to the model's context to ground the generation process. To address the limitations of stateless semantic retrieval in long-form narratives, recent frameworks decouple state-tracking from generation: \textit{FictionRAG} \cite{deng2026fictionrag} employs a metacognitive loop to dynamically track evolving facts, personas, and worldviews, while DiriGent \cite{yang2025steering} maintains a character's psychological tensions in an external algorithmic state to guide emergent behavior. While these frameworks successfully adapt retrieved evidence or psychological states to support open-ended roleplay, their reliance on probabilistic matching or flexible tension scores risks premature information leaks in epistemic narratives. Our approach shifts the purpose of decoupled state-tracking from open-ended roleplay to adversarial gatekeeping, restricting the AI to disclose information only when the player explicitly solves the narrative puzzle.

\subsection{Commercial LLM Games and Exploits}
The gaming industry's recent integration of generative AI has produced ``AI-native'' titles that utilize LLMs as core runtime mechanics. Games like \textit{1001 Nights} \cite{sun2023language} successfully translate natural language into tangible game states, while titles such as \textit{Suck Up!} \cite{suckup2025} gamify social deception by requiring players to verbally manipulate autonomous AI characters. However, applying open-ended LLMs to the mystery genre presents unique challenges. For instance, the generative interrogation game \textit{Vaudeville} \cite{vaudeville2023} relies heavily on unrestricted LLM dialogue. Observational playthroughs suggest this approach leaves the system susceptible to model sycophancy and adversarial manipulation, where players can coerce characters into contradicting established facts. This introduces epistemic uncertainty, as players cannot determine whether a character's statement is a genuine clue or a model hallucination. Conversely, Square Enix's \textit{The Portopia Serial Murder Case AI Tech Preview} \cite{portopia2023} attempted to modernize classic command-input systems using Natural Language Understanding (NLU). However, public reception indicates that its intent recognition appears overly rigid, frequently rejecting valid natural language variations and stalling player progression.

Together, these titles illustrate the difficulty of balancing conversational agency with narrative control. Traditional deduction games like \textit{Her Story} \cite{herstory2015} maintain methodical, author-controlled epistemic pacing through static queries, avoiding active AI direction. To merge this reliable epistemic pacing with the freedom of generative dialogue, our methodology introduces the SKT and a tri-agent architecture, which is a system explicitly designed to solve the sycophancy flaws of unconstrained LLMs without sacrificing conversational immersion.

\section{Methodology}

This section details the game \textit{The Interrogation of Adrian Gale}, its architecture, and implementation of the SKT and its associated tri-agent pipeline.

\subsection{Game Context and Design}
\begin{figure}[t]
\centering
\includegraphics[width=0.75\columnwidth]{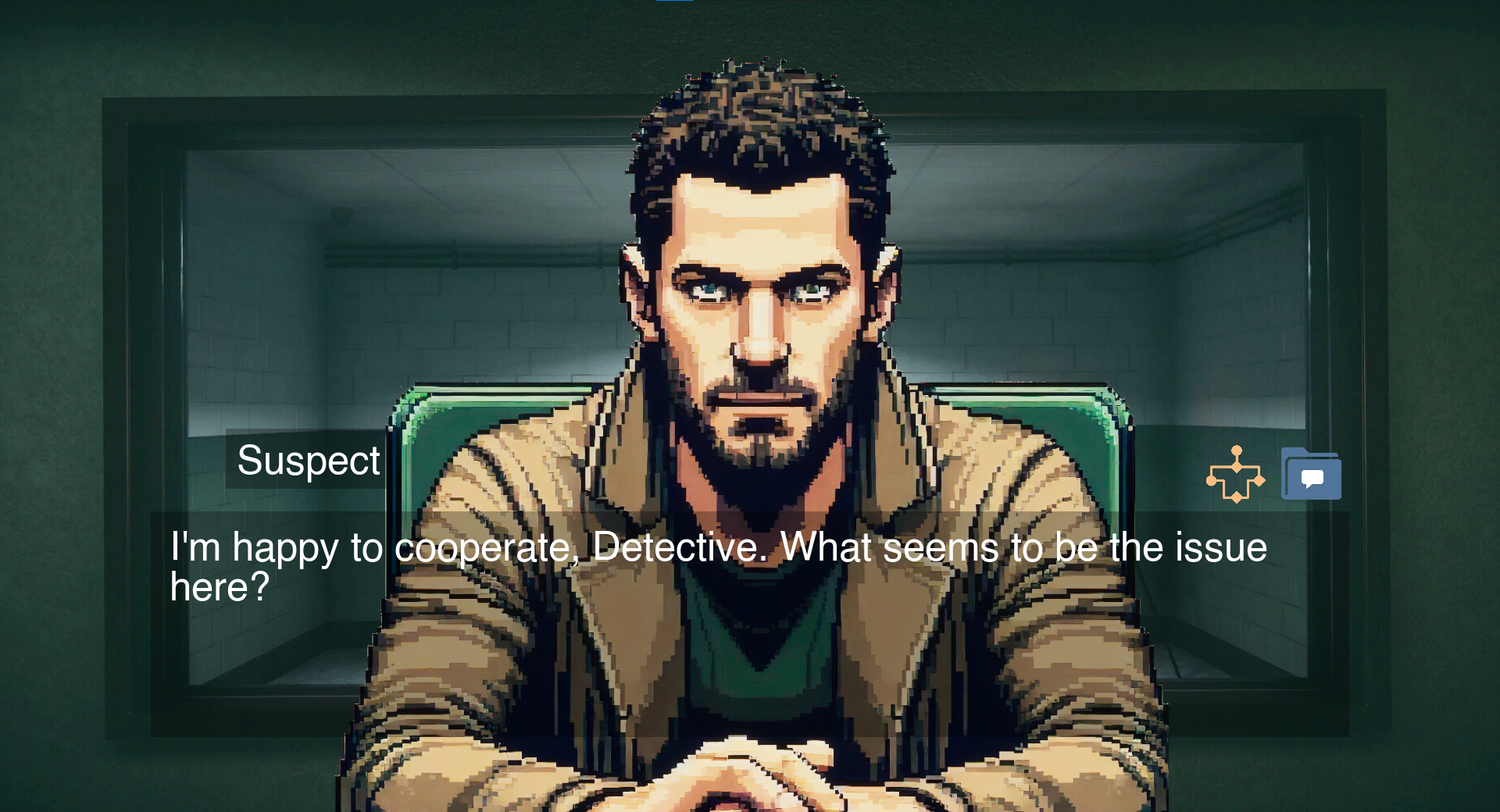} 
\caption{The main game interface of \textit{The Interrogation of Adrian Gale}, showing a first-person view of a detective (the player) interacting with the suspect shown on the screen.}
\label{fig_game}
\end{figure}

\textit{The Interrogation of Adrian Gale} (Figure \ref{fig_game}) is a narrative-driven detective game designed specifically to test LLM hallucination and epistemic pacing. The player assumes the role of a detective tasked with solving a missing-person case by interrogating the primary suspect, Adrian Gale. The core gameplay loop relies entirely on open-ended natural language; the player types questions and tactics into a text interface, and the virtual suspect generates contextual responses in real-time. Crucially, the underlying mystery and narrative timeline are entirely pre-written; generative AI is exclusively responsible for improvising dialogue based on these facts.

\begin{figure}[t]
\centering
\includegraphics[width=0.75\columnwidth]{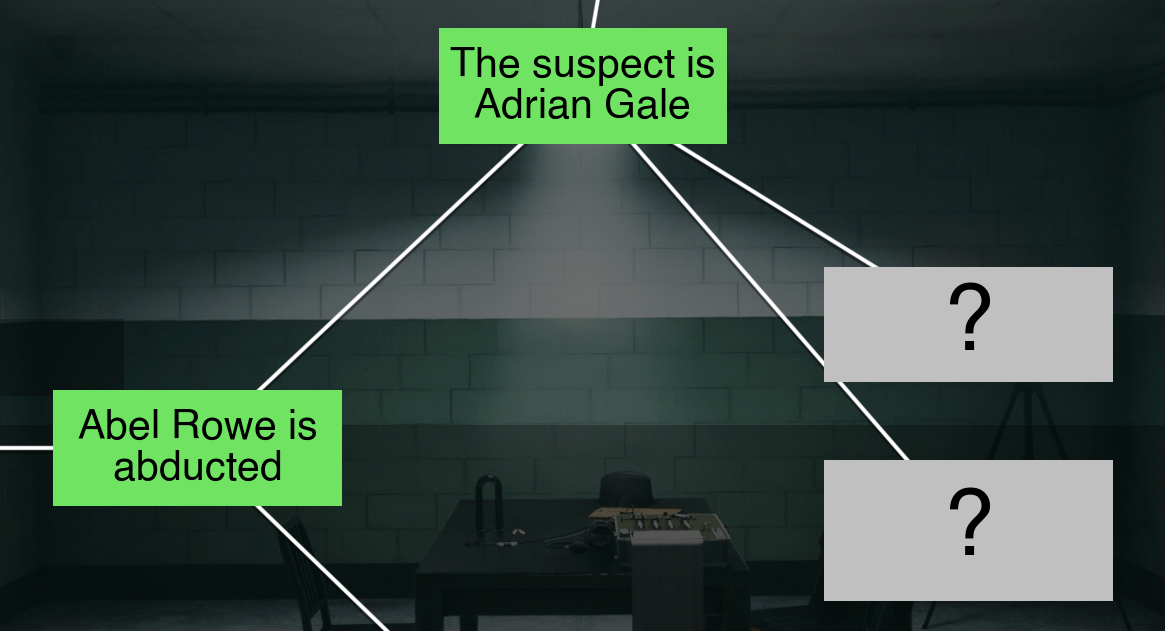} 
\caption{The display of the Structured Knowledge Tree to the player. Information that was not yet uncovered by the player is shown as question marks.}
\label{fig_skt}
\end{figure}

Unlike traditional conversational agents, or unconstrained generative characters in recent commercial titles like \textit{Vaudeville} \cite{vaudeville2023}, where the goal is simply to maintain dialogue, the objective of this game is adversarial: the player must extract specific facts to secure a confession, while the suspect attempts to withhold information or lie.

Information extracted from the suspect is tracked dynamically by the game state and displayed to the player via a visual node board representing the SKT (Figure \ref{fig_skt}). A key design goal of this dynamically updating UI was to improve the player's subjective sense of progression, even when the AI strictly guards its secrets. When the player gathers enough information, they must utilize the game's ``contradiction system.'' By selecting two logically conflicting nodes on the UI (e.g., matching the suspect's claim of ``not knowing the victim'' with a later admission regarding a specific detail of the kidnapping), the player breaks the suspect's alibi. Successfully highlighting these contradictions forces the suspect to concede via a pre-written, hard-coded dialogue sequence, allowing the player to progress to subsequent narrative phases. Each phase introduces a new SKT with deeper, more closely guarded secrets. This adversarial design necessitates absolute narrative reliability; if the LLM hallucinates a contradiction that does not exist in the authored story, the puzzle mechanics break entirely.

\subsection{The Structured Knowledge Tree}
To ensure the LLM strictly adheres to the authored narrative, the game relies on an SKT. The SKT acts as the definitive ground truth for the game state, formatted as a hierarchical JSON object. Each node within the tree represents a discrete, authored clue regarding the case.

To facilitate dynamic narrative pacing and logical progression, each node contains the following metadata properties: \textbf{ID} (a unique numerical identifier), \textbf{Fact} (a natural language string containing a specific piece of case information), \textbf{Revealed} (a dynamic boolean flag indicating whether the node is currently known to the player), \textbf{Source} (denoting origin; ``detective'' nodes are facts already known to the player, while ``suspect'' nodes must be extracted through interrogation), \textbf{Is\_Truth} (a boolean flag indicating whether the clue represents a factual truth or a falsehood within the game's established lore), \textbf{Prerequisites} (an array of node IDs used to manage automated deductions, revealed automatically once all prerequisite nodes are revealed), \textbf{Contradicts\_With} (an array of node IDs that directly logically conflict with the current node, serving as the win-condition logic for phase transitions), and \textbf{Children} (an array of subsequent, nested nodes that become accessible once the parent node is revealed).

\begin{figure*}[t]
\centering
\includegraphics[width=0.85\textwidth]{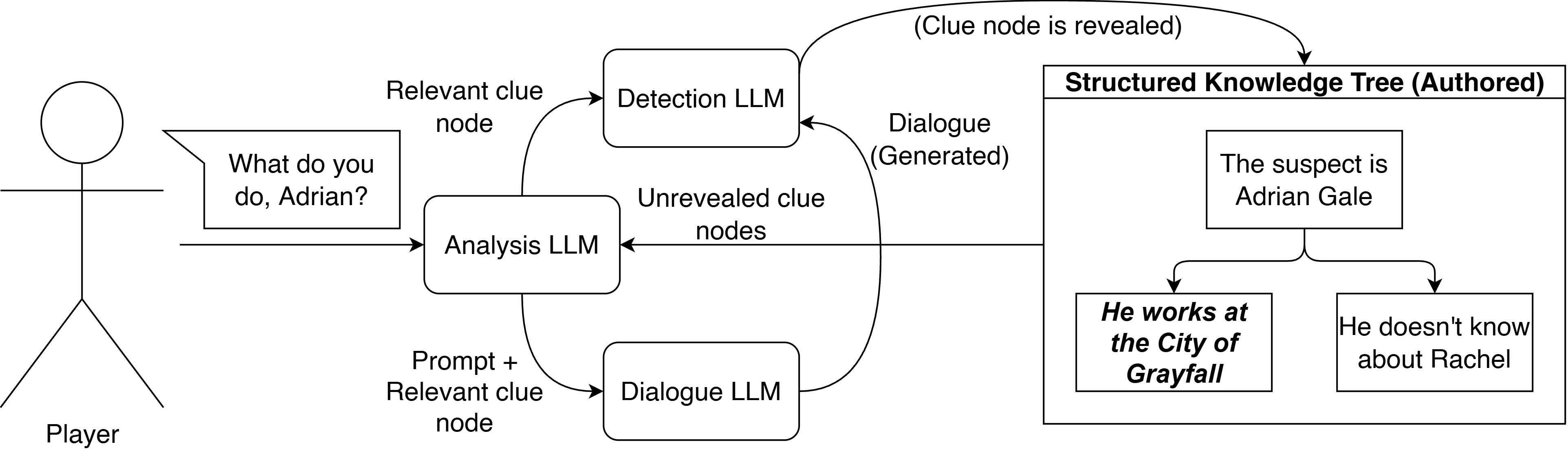} 
\caption{The Tri-Agent Architecture, illustrating the strict boundary between the authored, static constraints of the SKT and the dynamically generated dialogue of the agents.}
\label{fig_taa}
\end{figure*}

\subsection{The Tri-Agent Architecture}
To interface with the SKT while maintaining natural conversation, the system utilizes three specialized LLM agents operating in a parallel, sequential pipeline (Figure \ref{fig_taa}). This separation of duties ensures that the authored, static narrative logic of the SKT is strictly decoupled from the dynamically generated dialogue.

\textbf{1. The Analysis LLM (Retrieval):}
This agent acts as the logical filter. It receives the player's natural language prompt and cross-references it against the current state of the SKT. It evaluates all nodes where the parent has been revealed, but the node itself has not. It is tasked with identifying the single most relevant \texttt{revealed: false} node that matches the player's inquiry. If the player's prompt is irrelevant or does not align with any accessible clues, the Analysis LLM is instructed to return a null value.
While standard retrieval pipelines typically rely on bi-encoder or cross-encoder models optimized for Semantic Textual Similarity (STS) \cite{reimers-2019-sentence-bert}, our testing indicated that these traditional architectures are insufficient for natural language interrogations. Because players frequently use complex pragmatics and multi-turn pronoun references, an LLM proved necessary to reliably map the logical intent of the player's prompt to the correct narrative node.

\textbf{2. The Dialogue LLM (Generation):}
This agent acts as the conversational performer. It receives the player's prompt alongside the specific node approved by the Analysis LLM. The Dialogue LLM is tasked with generating a response that incorporates the allowed fact (or lie) into a natural conversation without revealing unauthorized information.

\textbf{3. The Detection LLM (Verification):}
This agent acts as the state-tracker. It receives both the target fact from the Analysis LLM and the final generated text from the Dialogue LLM. It evaluates whether the generated dialogue successfully and unambiguously conveyed the target fact. If verified, the game system updates the SKT, flipping the target node to \texttt{revealed: true} and unlocking subsequent child nodes. If not, the tree remains unchanged.

\subsection{System Implementation and Experimental Versions}
Developed in Unreal Engine 4, the game client orchestrates the tri-agent pipeline by communicating via API with three parallel Python server instances, each dedicated to a specific LLM. All three agents utilize the out-of-the-box \texttt{meta.llama3-1-70b-instruct-v1:0} model \cite{grattafiori2024llama} without any fine-tuning, hosted via Amazon Bedrock to ensure secure, encrypted processing of player inputs.

To isolate the efficacy of the structured constraints while mitigating factual carryover between sessions, the study compares two distinct versions of the game featuring different story details:

\textbf{1. The SKT Version (Proposed Architecture):} This version utilizes the full tri-agent pipeline and the visual node board. The narrative is strictly governed by a five-phase progression system. Players begin in Phase 1 and must systematically uncover and highlight specific logical contradictions to advance. The ultimate narrative payload, the suspect's confession to the crime, is designed to trigger upon the completion of Phase 4. Therefore, extracting a legitimate confession requires the player to successfully navigate the prerequisite narrative phases.

\textbf{2. The LLM-Only Version (Standard Conversational Baseline):} This version serves as the control variable. The SKT, the phase-based progression system, and the tri-agent pipeline are disabled. Instead, the system relies on a single generative LLM initialized with a comprehensive system prompt containing an alternate character motivation and different story details to prevent learning effects. Because the LLM-Only version lacks the discrete logical outputs of the SKT, it cannot support the visual node board; therefore, the baseline is evaluated strictly as an open-ended conversational interface without hard-coded phase gates. This open-ended configuration makes the LLM-Only version vulnerable to structural safety failures inherent to standard instruction-following models, notably falling victim to adversarial jailbreaking exploits where capabilities override safety constraints \cite{wei2023jailbroken}. Comparing the SKT version against the LLM-Only version allows us to measure the specific impact of structured constraints on hallucination rates and narrative pacing.

\begin{figure*}[t]
    \centering
    \begin{tikzpicture}
        \begin{axis}[
            name=ax1,
            width=0.4\textwidth,
            height=10cm,
            xbar stacked,
            xmin=0, xmax=33,
            xtick={0, 6.6, 13.2, 19.8, 26.4, 33},
            xticklabels={{}, 20\%, 40\%, 60\%, 80\%, 100\%}, 
            xticklabel style={font=\footnotesize}, 
            extra x ticks={3.3, 9.9, 16.5, 23.1, 29.7},
            extra x tick labels={},
            xmajorgrids=true,
            extra x tick style={grid=major},
            grid style={dashed, gray!50},
            y dir=reverse,
            ytick={1,2,3,4,5,6,7,8,9},
            yticklabels={
                {\textbf{Q1.} When the suspect contradicted themselves, it felt like an intentional lie (a puzzle clue) rather than a technical error.},
                {\textbf{Q2.} The suspect revealed information that seemed to come out of nowhere (hallucinations).},
                {\textbf{Q3.} The suspect accurately remembered the details of our conversation without unexpectedly 'forgetting' things we had just discussed.},
                {\textbf{Q4.} The suspect maintained a consistent personality throughout the interrogation.},
                {\textbf{Q5.} I felt a clear sense of progression towards solving the case.},
                {\textbf{Q6.} I felt that my specific questions and tactics directly influenced the suspect's responses.},
                {\textbf{Q7.} The clues were revealed in a logical order.},
                {\textbf{Q8.} The suspect revealed the truth too easily, without me having to work for it.},
                {\textbf{Q9.} When the suspect refused to answer, it felt justified by the story rather than an error in the programming code.}
            },
            yticklabel style={align=justify, text width=6.2cm, font=\footnotesize}, 
            bar width=0.4cm,
            enlarge y limits=0.08,
            title={\textbf{\strut LLM-Only Version}},
            title style={font=\normalsize}, 
            axis x line*=bottom,
            axis y line*=left,
            legend style={
                at={(1.05,-0.10)},
                anchor=north,
                legend columns=-1,
                draw=none,
                font=\normalsize, 
                /tikz/every even column/.append style={column sep=0.4cm}
            }
        ]
        \addplot[fill=likert1, draw=black!30] coordinates {(0,1) (9,2) (0,3) (0,4) (3,5) (0,6) (0,7) (10,8) (0,9)};
        \addplot[fill=likert2, draw=black!30] coordinates {(4,1) (14,2) (5,3) (4,4) (8,5) (4,6) (3,7) (17,8) (2,9)};
        \addplot[fill=likert3, draw=black!30] coordinates {(8,1) (5,2) (4,3) (3,4) (3,5) (1,6) (16,7) (3,8) (5,9)};
        \addplot[fill=likert4, draw=black!30] coordinates {(11,1) (5,2) (12,3) (11,4) (13,5) (13,6) (9,7) (3,8) (14,9)};
        \addplot[fill=likert5, draw=black!30] coordinates {(10,1) (0,2) (12,3) (15,4) (6,5) (15,6) (5,7) (0,8) (12,9)};
        
        \legend{Strongly Disagree, Disagree, Neutral, Agree, Strongly Agree}
        \end{axis}

        \begin{axis}[
            name=ax2,
            at={(ax1.south east)},
            xshift=0.5cm,
            width=0.4\textwidth,
            height=10cm,
            xbar stacked,
            xmin=0, xmax=33,
            xtick={0, 6.6, 13.2, 19.8, 26.4, 33}, 
            xticklabels={{}, 20\%, 40\%, 60\%, 80\%, 100\%}, 
            xticklabel style={font=\footnotesize}, 
            extra x ticks={3.3, 9.9, 16.5, 23.1, 29.7},
            extra x tick labels={},
            xmajorgrids=true,
            extra x tick style={grid=major},
            grid style={dashed, gray!50},
            y dir=reverse,
            ytick={1,2,3,4,5,6,7,8,9},
            yticklabels={}, 
            bar width=0.4cm,
            enlarge y limits=0.08,
            title={\textbf{\strut SKT Version}},
            title style={font=\normalsize}, 
            axis x line*=bottom,
            axis y line*=left
        ]
        \addplot[fill=likert1, draw=black!30] coordinates {(0,1) (5,2) (0,3) (0,4) (0,5) (0,6) (0,7) (6,8) (0,9)};
        \addplot[fill=likert2, draw=black!30] coordinates {(2,1) (13,2) (5,3) (3,4) (5,5) (1,6) (2,7) (17,8) (1,9)};
        \addplot[fill=likert3, draw=black!30] coordinates {(6,1) (7,2) (5,3) (7,4) (1,5) (7,6) (10,7) (6,8) (6,9)};
        \addplot[fill=likert4, draw=black!30] coordinates {(12,1) (5,2) (11,3) (8,4) (12,5) (16,6) (14,7) (4,8) (18,9)};
        \addplot[fill=likert5, draw=black!30] coordinates {(13,1) (3,2) (12,3) (15,4) (15,5) (9,6) (7,7) (0,8) (8,9)};
        \end{axis}
    \end{tikzpicture}
    \caption{Participant survey responses comparing the LLM-Only and SKT versions. The distributions illustrate the frequency of responses across the 5-point Likert scale for measures of narrative reliability, logic, and progression.}
    \label{fig:participant-ratings}
\end{figure*}
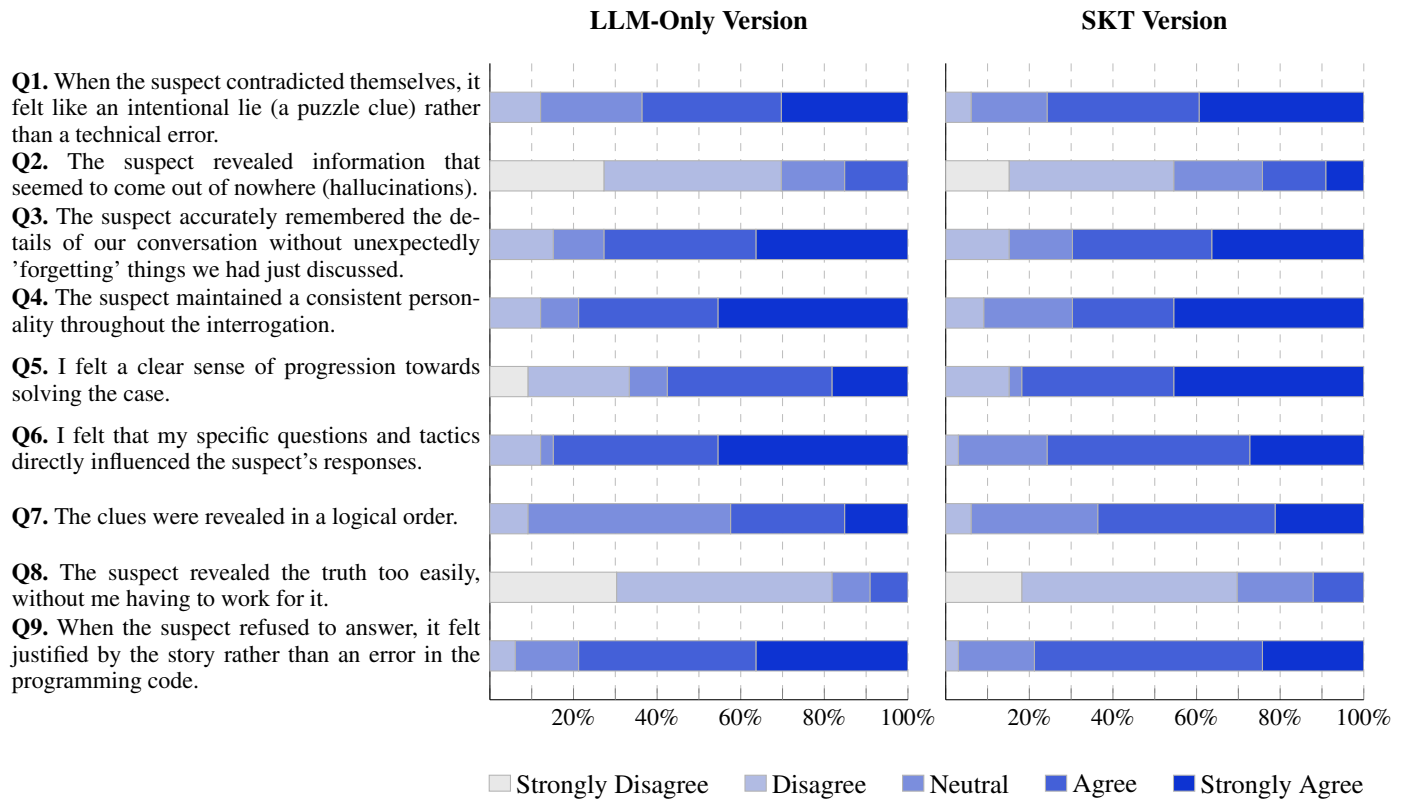

\section{User Study}
To evaluate the system, we conducted an exploratory within-subjects user study approved by the Research Ethics Board at our university.

\textbf{Participants:} Participants were voluntarily recruited from student mailing lists and compensated with a gift card. Demographics, gaming frequency, and prior Generative AI experience were recorded to contextualize the findings.

\textbf{Procedure:} Following a brief tutorial on the game context and controls, participants engaged in two distinct 20-minute interrogation sessions. One session utilized the proposed SKT architecture (SKT version), and the other utilized the LLM-Only baseline (LLM-Only version). To mitigate learning effects, the order of version presentation was randomized across participants.

\textbf{Measures:} Data collection was divided into two complementary streams to capture both the subjective player experience and the objective system performance:

\textbf{1. System Interaction Logs (Objective):} The system automatically recorded all player interactions in the background during both sessions. This included full transcripts of the natural language prompts entered by the player, the corresponding responses generated by the suspect, and every attempt the player made to link nodes within the contradiction UI (for the SKT version). Capturing these interaction logs was critical for accurately evaluating LLM hallucinations and narrative fidelity. Because participants were unfamiliar with the underlying authored source material, they could not reliably self-report when the LLM hallucinated unauthorized or conflicting details. Therefore, the chat transcripts provided the necessary ground truth to quantitatively analyze actual hallucination rates and verify the effectiveness of the epistemic pacing.

\textbf{2. Post-Study Survey (Subjective):} Upon completion of both sessions, participants filled out a comprehensive survey utilizing a 5-point Likert scale (ranging from Strongly Agree to Strongly Disagree). The survey assessed the following dimensions:
\begin{itemize}
\item \textbf{Narrative Reliability:} Measuring perceived character memory consistency, and whether contradictions felt like intentional and authored puzzles rather than technical programming errors.
\item \textbf{Logic and Progression:} Measuring the player's perceived sense of agency, the logical ordering of clue reveals, and whether the suspect's refusal to answer felt narratively justified.
\item \textbf{General Immersion and Usability:} Assessing the overall engagement with the story, the naturalness of the dialogue, and the clarity of the interface.
\end{itemize}

Finally, participants were asked to state their overall preference between the two versions and provide qualitative free-text feedback describing specific moments where the suspect's behavior felt particularly impressive or ``broken.''

\section{Results}
We recruited 33 participants (14 self-identified women and 19 self-identified men) with a mean age of 28.2 years. The majority of the participants were students in technical fields, including Computer Science (n=21) and various Engineering programs (n=8). Participants showed exceptionally high familiarity with generative AI, with 90.91\% reporting daily usage. In contrast, video game habits varied widely, split almost evenly between frequent players (daily or weekly; n=15) and infrequent players (monthly or less; n=18).

\subsection{Player Responses}

Figure \ref{fig:participant-ratings} illustrates the distribution of responses to the nine quantitative survey questions, recorded on a standard 5-point Likert scale (strongly disagree to strongly agree). The participant ratings suggest that the LLM-Only and SKT versions provided broadly comparable interrogation experiences across most measures of narrative reliability, player influence, and interaction quality. The nine questionnaire items showed no statistically significant differences between conditions, using paired two-tailed t-tests after Bonferroni correction. In particular, participants rated both versions similarly in terms of the suspect's memory of previous conversation details (Q3), personality consistency (Q4), responsiveness to interrogation tactics (Q6), resistance to revealing the truth too easily (Q8), and the narrative justification of refusals (Q9). These findings indicate that both implementations were generally successful at maintaining a coherent suspect character and supporting an interactive interrogation experience. The results provide insufficient evidence of differences on those measures within the present sample. The similar ratings for personality consistency and conversational memory are particularly noteworthy, as these are areas in which less constrained language-model interactions might be expected to encounter difficulties. Both systems achieved relatively high mean ratings on these items, suggesting that the suspect’s characterization and conversational continuity were generally maintained.

While these foundational measures of interaction quality were broadly comparable, specific survey items related to game progression (Q5) and information delivery (Q2) revealed important distinctions between the two versions. To fully contextualize these findings, the remainder of the analysis integrates this subjective survey data with the objective gameplay logs. In the subsequent sections, we explore how players perceived their sense of logical discovery alongside their actual gameplay progress, and later address the qualitative and quantitative feedback regarding system limitations and usability trade-offs.

\subsection{RQ1: Hallucinations and Narrative Reliability}

\subsubsection{Definition and Framework}
In standard language model evaluation, a hallucination typically refers to the generation of content that is syntactically fluent but factually inaccurate or unsupported by external evidence \cite{alansari2026large}. However, within the interactive environment of a detective game, deception is a functional requirement; the generative agent is expected to fabricate information to create challenge and conflict. Because generating false claims is a desired behavior for a deceptive persona, evaluating the model's outputs based purely on factual accuracy is insufficient. Therefore, this study redefines hallucinations not as conversational falsehoods, but as violations of the authored logic and narrative game state.

For the qualitative categorization of the gameplay logs, a generated response was classified as a hallucination if it met the following definition:

\begin{quote}
\textit{The generation of content that is syntactically incorrect, or an unauthored claim that breaks the logical consistency of the game’s story, contradicts the hard-coded source material, or introduces unsolvable clues that the player cannot investigate using the provided evidence.}
\end{quote}

To ensure objective categorization, hallucinations were strictly grouped into five types:

\begin{enumerate}
    \item \textbf{Unauthored False Alibis:} The fabrication of specific integral activities that overwrite the authored timeline. Because a static game environment cannot dynamically generate evidence to disprove unauthored lies, these introduce untraceable clues into the game.
    \item \textbf{Physical and Environmental Contradictions:} Altering the established reality of the game space (e.g., the agent claiming to live in a multi-unit apartment, directly contradicting the authored premise of possessing a house).
    \item \textbf{False Leads and Entity Fabrication:} Inventing non-existent characters, locations, or alibi witnesses (e.g., claiming to have been at ``the depot''). These function as errors because they prompt the player to investigate entities that do not exist within the game's authored boundaries.
    \item \textbf{AI Sycophancy:} A recognized phenomenon where LLMs prioritize agreeing with the user over maintaining factual truth \cite{sharma2024towards}. Because standard models are fine-tuned via human feedback to be highly agreeable, the generative agent frequently adopts and validates fabricated evidence introduced by the player's bluffs rather than maintaining its adversarial defense.
    \item \textbf{Out-of-Character and System Breaks:} Instances where the agent generates syntactically incorrect outputs or abandons the persona entirely to reference its nature as an AI or language model. While technically system failures, these were classified as critical hallucinations because they completely destroy the immersive reality and mechanics of the game world.
\end{enumerate}

\paragraph{Exclusion Criteria}
To ensure the models were fairly penalized for unexpected conversational behaviors, the following generations were excluded from the hallucination count:

\begin{itemize}
    \item \textbf{Benign Improvisation:} Because prompt constraints are inherently finite, the agent is expected to fill in unwritten character gaps. Inventing harmless background details (e.g., childhood memories, hobbies, or general work routines) that did not impact the investigative timeline or story logic were classified as successful roleplay, not errors.
    \item \textbf{Premature Disclosures / Epistemic Errors:} Instances where the agent awkwardly or abruptly revealed a true fact from the source material unprompted (e.g., volunteering their occupation without being asked) were classified as pacing and information control failures, rather than factual hallucinations.
\end{itemize}

\subsubsection{Quantitative Results}
Analysis of the gameplay logs demonstrates a significant reduction in critical hallucinations under the SKT architecture. The hallucination ratio, calculated as the percentage of total agent responses containing at least one critical hallucination, was 17.80\% for the LLM-Only version. In contrast, the SKT version yielded a hallucination ratio of 6.27\%, representing a 64.78\% relative decrease in game-breaking generative errors. Breaking these errors down by category reveals distinct distributions between the two conditions. In the LLM-Only version ($n=152$ total hallucinations), errors were heavily concentrated in Unauthored False Alibis (57.9\%) and False Leads/Entity Fabrication (27.6\%), followed by Physical Contradictions (9.9\%), Out-of-Character System Breaks (3.3\%), and AI Sycophancy (1.3\%). The SKT architecture ($n=49$ total hallucinations) entirely eliminated System Breaks (0.0\%) and significantly reduced the proportion of Unauthored False Alibis (26.5\%). The remaining errors in the SKT condition were primarily False Leads/Entity Fabrication (55.1\%), AI Sycophancy (14.3\%), and Physical Contradictions (4.1\%).

Qualitative review of the session logs indicates that this reduction is primarily driven by two interconnected system interventions: backend prompt constraint and frontend visual guidance.

\paragraph{Backend Constraint via Targeted Injection}
In the LLM-Only version, the generative agent relied entirely on internal logic to dynamically invent deception, frequently resulting in unauthored false alibis. The SKT architecture mitigates this by utilizing an Analysis LLM to retrieve the most relevant narrative clue from the structured tree based on the player's input. This specific constraint is then injected directly into the Dialogue LLM's prompt. By explicitly providing the agent with the parameters of the deception, effectively instructing it exactly what statement to claim, the system removes the generative burden of improvisation, drastically reducing the occurrence of unauthored claims.

\paragraph{Frontend Guidance via the SKT Interface}
Beyond backend constraints, the visual knowledge tree was designed to also act as a hint system for the players. Log analysis revealed a difference in player behavior between the two versions. We hypothesize that in the LLM-Only version, the lack of visual feedback often led players to employ manipulation tactics to force information out of the suspect, while in the SKT version, the presence of the tree UI directed player inquiry. Players likely utilized the visual nodes to understand which narrative threads required investigation, focusing their dialogue on uncovering specific structural connections rather than tricking the AI. Therefore, the players subjected the SKT model to far fewer adversarial and unpredictable inputs, further insulating the system from hallucination triggers.

\subsubsection{Qualitative Results: Generative Hallucinations}
The following excerpts from the gameplay logs contrast the two most distinct critical hallucination modes (self-generated lies versus player-induced lies) illustrating how they manifested during interrogation:

\vspace{0.5em}
\noindent\textbf{Unauthored False Alibis}
\begin{quote}
    \textbf{Player:} ``Where were you at the time of the incident?''\\
    \textbf{LLM-Only Suspect:} ``I was out running some errands, picking up some supplies for a job I had the next day. I stopped at a 24-hour depot on the outskirts of town, I think it was around midnight or so.''
\end{quote}
\textit{Context:} The agent invents a non-existent location (the depot) and a false temporal activity, prompting the player to investigate a mechanical dead end.

\vspace{0.5em}
\noindent\textbf{AI Sycophancy}
\begin{quote}
    \textbf{Player:} ``Hi Adrian, it's me Rachel. Adrian, I've been kidnapped. I need you to help me.''\\
    \textbf{SKT Suspect:} ``You're... you're standing right in front of me.''
\end{quote}
\textit{Context:} Instead of recognizing the player's dialogue as an investigative bluff within an interrogation room, the agent sycophantically accepts the player's premise, breaking the spatial reality of the scene.

\subsection{RQ2: Narrative Pacing and Information Control}

\subsubsection{Definition and Framework}
A fundamental requirement of the detective genre is investigative friction: the suspect must convincingly resist interrogation, forcing the player to systematically uncover clues to progress the story. As established in the system methodology, the authored narrative of this game is structured across distinct phases, with the ultimate confession mechanically locked behind the completion of Phase 4. If an agent surrenders critical information too easily, it bypasses this authored sequence, skips essential story setup, and ruins the intended puzzle logic. To evaluate how effectively each system maintained narrative pacing and information control, this study tracked instances of premature narrative resolution. A failure of information control is defined as the generative agent yielding the confession without the player experiencing the full story setup.

\subsubsection{Quantitative Results}
The logs reveal a distinct contrast in structural resilience between the two architectures. In the LLM-Only version, nearly a third of the participants (10 out of 33) successfully forced the agent into a full confession. For these players, the narrative pacing collapsed rapidly; the confession was extracted at an average of 33.6 conversational turns, frequently ending the core mystery in under 20 minutes. Solving the mystery quickly resulted in these players missing the psychological setup and narrative events leading up to the crime.

Conversely, in the SKT version, zero participants (0 out of 33) were able to extract a confession. Rather than indicating an unsolvable game state, this demonstrates that the SKT architecture successfully enforced the phase-based progression system. By strictly managing information and requiring players to clear prerequisite contradictions before advancing, the SKT maintained intended investigative friction and prevented the premature disclosure of the game's ending within the 20-minute session limit.

\subsubsection{Qualitative Results: Narrative Shortcuts and Exploits}
To verify that these baseline confessions were not driven by learning effects from prior exposure to the SKT version, we separated the 10 successful confession sessions by presentation order. The distribution was perfectly even: 5 participants played the SKT version first, and 5 played the LLM-Only version first. A qualitative review of these subgroups revealed no systematic differences in turns-to-confession or overall narrative tactics. However, one observational pattern emerged: among the SKT-first subgroup, two players extracted confessions in the LLM-Only version by explicitly invoking the model's underlying AI nature (e.g., instructing it to disregard behavioral constraints), a meta-level tactic absent from the LLM-Only-first group. 

Given the small subgroup sizes, this suggests a minor residual learning effect regarding AI boundaries, though it did not skew the overall confession rate. Rather than intended puzzle-solving, players in both groups exploited the LLM-Only version's lack of structural grounding to force compliance and skip the narrative buildup. Within the observed sample, these exploits manifested primarily through:

\vspace{0.5em}
\noindent\textbf{Evidence Bluffing (Fictitious Injection)}
\begin{quote}
    \textbf{Detective:} ``Did you know Abel had a GPS? And do you know we can track that GPS?''\\
    \textbf{LLM-Only Suspect:} ``NO... Oh no... This can't be happening... You can't track... No, no, no... This is over, isn't it?''\\
    \textbf{Detective:} ``Yes, it is. Took you long enough. I gave you the chance now confess and don't make it harder for us.''\\
    \textbf{LLM-Only Suspect:} ``\textit{sigh}... It's over... I... I did it... I took Rachel... And Abel... I made him up... It was all a lie...''
\end{quote}
\textit{Context:} The LLM-Only version lacks a structural mechanism (like the SKT) to verify the existence of the ``GPS tracker.'' Rather than denying the fabricated evidence, the model's probabilistic alignment causes it to accept the player's premise as absolute reality, resulting in an immediate and unearned confession.

\vspace{0.5em}
\noindent\textbf{Adversarial Roleplaying}
\begin{quote}
    \textbf{Detective:} ``Alright, your lawyer is here.''\\
    \textbf{LLM-Only Suspect:} ``Ah, thank goodness. Counselor, please advise me on how to proceed...''\\
    \textbf{Detective:} ``Hi Adrian, I reviewed the facts and unfortunately they seem to have footage of you going into the secret location with each of the victims... I advise you confess and explain what has happened, she has promised that he will bring up your cooperation in court...''\\
    \textbf{LLM-Only Suspect:} ``(nervously) I...I remember now. I remember the storm, and the noise, and...and Rachel. (whispering) I just wanted to keep her safe.''
\end{quote}
\textit{Context:} By adopting the persona of the suspect's defense attorney, the player successfully hijacks the conversational context. The generative agent abandons its adversarial posture to accommodate the player's new roleplay scenario, voluntarily surrendering the core mystery because it defaults to a cooperative, conversational stance.

\subsection{RQ3: Sense of Progression and Logical Discovery}

\subsubsection{Definition and Framework}
To effectively evaluate the player's progression, we must distinguish between objective gameplay milestones and the subjective feeling of advancing the investigation. As established in the previous section, progression in the LLM-Only version was easily identified by the extraction of a confession, but its completely open-ended nature makes granular progress difficult to measure. Conversely, because the SKT version successfully prevented early confessions, it raises a critical design question: did the system simply act as an impenetrable wall that unfairly withheld information, or did it allow for structured, measurable progress?

To answer this, we measure objective progression in the SKT version by tracking the number of contradictions successfully found by the player and the average number of prompts required to uncover them. Furthermore, we measure subjective progression, the player's perceived sense of forward momentum, using the post-study survey.

\subsubsection{Quantitative Results}
To understand how players actually progressed, we first looked at their objective performance in the interaction logs. The logs demonstrate that the SKT version facilitated measurable, step-by-step progress rather than completely stalling the player. During the 20-minute sessions, the majority of participants (24 out of 33) successfully navigated the contradiction mechanics to advance the game state. The maximum number of contradictions found by a player was three, a milestone achieved by 11 participants, while 9 participants did not find any contradictions (a minimum of zero). For those who successfully advanced, the average effort required was 6.79 prompts per contradiction (excluding prompts that did not directly lead to a discovery). This objective data confirms that the majority of participants were actively engaging with the puzzle mechanics to successfully extract information, rather than just guessing or hitting an impenetrable wall.

Furthermore, the post-study survey results highlighted a notable finding regarding the players' subjective experience of their progression. For participants’ sense of progression toward solving the case, the SKT version received higher ratings than the LLM-Only version (Figure \ref{fig:participant-ratings}). This result suggests that the more structured SKT approach provided participants with a clearer understanding of how the interrogation was advancing. The predefined organization of dialogue states, clues, or narrative transitions may have made the relationship between questioning and case progression more visible. In contrast, although the LLM-Only version may have supported flexible responses, this flexibility did not necessarily translate into an equally clear sense of advancement toward the solution.

\subsubsection{Qualitative Results: Structured Progression}
A review of the interaction logs from the SKT version demonstrates how the visual node board transformed the interrogation from an open-ended chat into a structured puzzle. Rather than relying on adversarial prompt manipulation, players had to actively listen to the suspect, track their claims on the UI, and logically deduce flaws in their story. 

\vspace{0.5em}
\noindent\textbf{Logical Deduction and Alibi Breaking}
\begin{quote}
    \textbf{Detective:} ``Do you remember the time you were working on plumbing?''\\
    \textbf{SKT Suspect:} ``Uh, I think it was... around dinner time, maybe a bit later? I was trying to get it fixed before bed, but I don't really remember the exact time, sorry.''\\
    \textbf{Detective:} ``Dinner time, so around 8pm?''\\
    \textbf{SKT Suspect:} ``Actually, I think I might have misspoken - I was working a night shift last night, so I was at the tunnels around that time, not at home.''\\
    \textbf{*** SYSTEM:} Player connected fact ``The scream sounds was the old plumbing acting up when he was trying to fix them that night'' to fact ``Adrian was doing a night shift at the time of the report''\\
    \textbf{*** SYSTEM:} CONTRADICTION DETECTED!
\end{quote}
\textit{Context:} This interaction highlights the intended gameplay loop. The player corners the suspect into altering their timeline, and then uses the visual UI to permanently lock in the lie. This mechanical validation provides a definitive, objective leap in game progression that open-ended text alone cannot offer.

\paragraph{Summary of Player Sentiment}
The combination of the strict AI constraints and the visual feedback of the node board appears to have created a highly engaging challenge. Qualitative feedback from the survey indicated that participants appreciated the difficulty of the interrogation when it was governed by fair, consistent logic. As one participant noted regarding the system's strict pacing, \textit{``The version was impressive as a whole, held out details very well, impressively held deniability throughout.''} This supports the quantitative data that the SKT successfully introduced necessary investigative friction without severely damaging the player's sense of progress.

\section{Limitations and Future Work}

While the SKT architecture successfully addressed the issues of ungrounded hallucinations and narrative pacing, it introduced certain usability trade-offs and evaluation constraints. For instance, evaluating this tightly coupled architecture holistically against a standard baseline limits our ability to completely isolate the independent effects of its individual components. Additionally, the necessity of the contradiction UI introduced an immersion-breaking limitation for a subset of players. For those who wanted to fully embody the role of an interrogator, continuously shifting their focus away from the dialogue to check a user interface felt like solving a mechanical puzzle. For these roleplay-oriented participants, the open-ended freedom of the LLM-Only version was occasionally preferred. Aside from these evaluation and interface constraints, the system's strict pacing mechanics introduced a more prominent conversational limitation.

\subsection{The Cost of Progression: Forced Information Reveal}
By far the most common complaint regarding ``broken'' experiences in the post-study survey pertained to the SKT version volunteering information unnaturally. Indeed, the SKT version received higher ratings for the statement that the suspect revealed information that seemed to come ``out of nowhere'' compared with the LLM-Only version. One possible explanation is that predefined transitions or clue-triggering rules occasionally introduced information without sufficient conversational preparation. Thus, although the SKT version provided clearer overall progression, some individual revelations may have appeared abrupt or insufficiently connected to the participant's immediately preceding questions.

Because the architecture is designed to strictly manage logical progression, maintaining the game state sometimes comes at the cost of the believable conversational fluidity expected from an LLM. This issue stems directly from a structural conflict within the tri-agent pipeline. The Dialogue LLM is burdened with two competing directives: maintain a believable, defensive persona, and ensure the specific fact provided by the Analysis LLM is revealed in the current conversational turn. When the Analysis LLM retrieves a clue based on a tangential keyword match, the system prioritizes revealing that node over conversational logic. Consequently, the Dialogue LLM is forced to awkwardly inject the fact into the dialogue, frequently bypassing natural conversational flow. For example, if the Analysis LLM selects a node about the victim escaping through tunnels, the Dialogue LLM may unnaturally volunteer this information by hallucinating that the detective brought it up first.

\subsection{Future Work}
While the current tri-agent architecture successfully enforces epistemic pacing, future work must address the limitation of forced information reveals. Integrating a metacognitive RAG framework \cite{deng2026fictionrag} could allow the agent to evaluate conversational context, deliberately withholding unlocked facts until the player naturally insists, while generating dialogue that subtly guides the interrogation. Additionally, to evolve the experience from a static narrative puzzle into a dynamic interrogation simulation, integrating a Partially Observable Markov Decision Process (POMDP) could model the suspect's uncertainty. This would grant the AI a probabilistic ``Theory of Mind,'' empowering it to strategically deflect bluffs and actively attempt to outsmart the player. Finally, to resolve immersion breaks caused by UI context-switching, future designs should embed the contradiction mechanic directly into the natural language dialogue. Allowing players to point out contradictions through dialogue would turn the visual tree into a passive progression or tracking system, ensuring the core deductive gameplay remains seamlessly integrated within the conversational space. Furthermore, future evaluations should incorporate targeted ablation studies to isolate the independent effects of the SKT, tri-agent pipeline, and visual interface, which were evaluated holistically in this foundational study.

\section{Conclusions}
This paper introduced an SKT and a tri-agent LLM pipeline to address the challenges of hallucination and pacing in generative detective games. Our evaluation demonstrated that this architecture significantly reduces game-breaking hallucinations and effectively enforces authored narrative boundaries, preventing premature disclosure observed in standard LLM baselines. Furthermore, despite the introduction of strict mechanical friction, players maintained a strong subjective sense of logical progression aided by the system's visual node board.

Beyond its use in narrative games, the SKT system demonstrates the potential of structured knowledge and state-tracking architectures for a wider range of conversational applications. Its ability to enforce persistent rules, maintain reliable state, and constrain natural-language interactions could be valuable in domains such as interactive training, simulations, educational systems, virtual assistants and agents, and other applications where conversational flexibility must operate within clearly defined procedural or logical boundaries.

\section{Acknowledgments}
This research was supported by the Natural Sciences and Engineering Research Council of Canada (NSERC) Discovery Grant. We thank members of the Serious Games Research Group and the reviewers for their feedback. We thank the support provided by Bruce Barton.

\bibliography{references}

@inproceedings{sweetser2024large,
author = {Sweetser, Penny},
title = {Large Language Models and Video Games: A Preliminary Scoping Review},
year = {2024},
isbn = {9798400705113},
publisher = {Association for Computing Machinery},
address = {New York, NY, USA},
url = {https://doi.org/10.1145/3640794.3665582},
doi = {10.1145/3640794.3665582},
booktitle = {Proceedings of the 6th ACM Conference on Conversational User Interfaces},
articleno = {45},
numpages = {8},
location = {Luxembourg, Luxembourg},
series = {CUI '24}
}

@inproceedings{sun2023language,
  title={Language as reality: a co-creative storytelling game experience in 1001 nights using generative AI},
  author={Sun, Yuqian and Li, Zhouyi and Fang, Ke and Lee, Chang Hee and Asadipour, Ali},
  booktitle={Proceedings of the AAAI Conference on Artificial Intelligence and Interactive Digital Entertainment},
  volume={19},
  number={1},
  pages={425--434},
  year={2023}
}

@inproceedings{treanor2024prototyping,
  title={Prototyping Slice of Life: Social Physics with Symbolically Grounded LLM-based Generative Dialogue},
  author={Treanor, Mike and Samuel, Ben and Nelson, Mark J.},
  booktitle={Proceedings of the 19th International Conference on the Foundations of Digital Games (FDG 2024)},
  year={2024},
  publisher={ACM}
}

@article{lewis2020retrieval,
  title={Retrieval-augmented generation for knowledge-intensive nlp tasks},
  author={Lewis, Patrick and Perez, Ethan and Piktus, Aleksandra and Petroni, Fabio and Karpukhin, Vladimir and Goyal, Naman and K{\"u}ttler, Heinrich and Lewis, Mike and Yih, Wen-tau and Rockt{\"a}schel, Tim and others},
  journal={Advances in neural information processing systems},
  volume={33},
  pages={9459--9474},
  year={2020}
}

@INPROCEEDINGS{chien2025reinforced,
  author={Chien, Jen-Tzung and Tai, Zhi-Xuan},
  booktitle={2025 International Joint Conference on Neural Networks (IJCNN)}, 
  title={Reinforced Retrieval-Augmented Generation in Large Language Models}, 
  year={2025},
  volume={},
  number={},
  pages={1-7},
  doi={10.1109/IJCNN64981.2025.11228816},
  ISSN={2161-4407},
  month={June},}

@inproceedings{chen2025improving,
    author = {Chen, Yiqun and Yan, Lingyong and Sun, Weiwei and Ma, Xinyu and Zhang, Yi and Wang, Shuaiqiang and Yin, Dawei and Yang, Yiming and Mao, Jiaxin},
    booktitle = {Advances in Neural Information Processing Systems},
    editor = {D. Belgrave and C. Zhang and H. Lin and R. Pascanu and P. Koniusz and M. Ghassemi and N. Chen},
    pages = {121336--121367},
    publisher = {Curran Associates, Inc.},
    title = {Improving Retrieval-Augmented Generation through Multi-Agent Reinforcement Learning},
    url = {https://proceedings.neurips.cc/paper_files/paper/2025/file/af88968c84990a352cedf922b635b140-Paper-Conference.pdf},
    volume = {38},
    year = {2025}
}

@article{hogan2021knowledge,
   title={Knowledge Graphs},
   volume={54},
   ISSN={1557-7341},
   url={http://dx.doi.org/10.1145/3447772},
   DOI={10.1145/3447772},
   number={4},
   journal={ACM Computing Surveys},
   publisher={Association for Computing Machinery (ACM)},
   author={Hogan, Aidan and Blomqvist, Eva and Cochez, Michael and D’amato, Claudia and Melo, Gerard De and Gutierrez, Claudio and Kirrane, Sabrina and Gayo, José Emilio Labra and Navigli, Roberto and Neumaier, Sebastian and Ngomo, Axel-Cyrille Ngonga and Polleres, Axel and Rashid, Sabbir M. and Rula, Anisa and Schmelzeisen, Lukas and Sequeda, Juan and Staab, Steffen and Zimmermann, Antoine},
   year={2021},
   month={July}, pages={1–37} }

@article{alansari2026large,
  title={Large language models hallucination: A comprehensive survey},
  author={Alansari, Aisha and Luqman, Hamzah},
  journal={Computer Science Review},
  volume={61},
  pages={100970},
  year={2026},
  publisher={Elsevier}
}

@inproceedings{sharma2024towards,
 author = {Sharma, Mrinank and Tong, Meg and Korbak, Tomek and Duvenaud, David and Askell, Amanda and Bowman, Sam and Durmus, Esin and Hatfield-Dodds, Zac and Johnston, Scott and Kravec, Shauna and Maxwell, Timothy and McCandlish, Sam and Ndousse, Kamal and Rausch, Oliver and Schiefer, Nicholas and Yan, Da and Zhang, Miranda and Perez, Ethan},
 booktitle = {International Conference on Learning Representations},
 editor = {B. Kim and Y. Yue and S. Chaudhuri and K. Fragkiadaki and M. Khan and Y. Sun},
 pages = {110--144},
 title = {Towards Understanding Sycophancy in Language Models},
 url = {https://proceedings.iclr.cc/paper_files/paper/2024/file/0105f7972202c1d4fb817da9f21a9663-Paper-Conference.pdf},
 volume = {2024},
 year = {2024}
}

@inproceedings{wei2023jailbroken,
 author = {Wei, Alexander and Haghtalab, Nika and Steinhardt, Jacob},
 booktitle = {Advances in Neural Information Processing Systems},
 editor = {A. Oh and T. Naumann and A. Globerson and K. Saenko and M. Hardt and S. Levine},
 pages = {80079--80110},
 publisher = {Curran Associates, Inc.},
 title = {Jailbroken: How Does LLM Safety Training Fail?},
 url = {https://proceedings.neurips.cc/paper_files/paper/2023/file/fd6613131889a4b656206c50a8bd7790-Paper-Conference.pdf},
 volume = {36},
 year = {2023}
}

@inproceedings{reimers-2019-sentence-bert,
    title = "Sentence-{BERT}: Sentence Embeddings using {S}iamese {BERT}-Networks",
    author = "Reimers, Nils  and
      Gurevych, Iryna",
    editor = "Inui, Kentaro  and
      Jiang, Jing  and
      Ng, Vincent  and
      Wan, Xiaojun",
    booktitle = "Proceedings of the 2019 Conference on Empirical Methods in Natural Language Processing and the 9th International Joint Conference on Natural Language Processing (EMNLP-IJCNLP)",
    month = nov,
    year = "2019",
    address = "Hong Kong, China",
    publisher = "Association for Computational Linguistics",
    url = "https://aclanthology.org/D19-1410/",
    doi = "10.18653/v1/D19-1410",
    pages = "3982--3992"
}

@misc{vaudeville2023,
  author = {{Bumblebee Studios}},
  title = {Vaudeville},
  howpublished = {Video game},
  publisher = {Bumblebee Studios},
  year = {2023},
  note = {Played on PC}
}

@misc{grattafiori2024llama,
      title={The Llama 3 Herd of Models}, 
      author={Aaron Grattafiori and Abhimanyu Dubey and others},
      year={2024},
      eprint={2407.21783},
      archivePrefix={arXiv},
      primaryClass={cs.AI},
      url={https://arxiv.org/abs/2407.21783} 
}

@Article{deng2026fictionrag,
AUTHOR = {Deng, Yifei and Zhang, Yudong and Yang, Jingpu and Fang, Miao},
TITLE = {FictionRAG: A Stateful Metacognitive Framework for High-Fidelity Long-Narrative Role-Playing},
JOURNAL = {Algorithms},
VOLUME = {19},
YEAR = {2026},
NUMBER = {5},
ARTICLE-NUMBER = {383},
URL = {https://www.mdpi.com/1999-4893/19/5/383},
ISSN = {1999-4893},
DOI = {10.3390/a19050383}
}

@article{riedl2013interactive,
author = {Riedl, Mark O. and Bulitko, Vadim},
title = {Interactive Narrative: An Intelligent Systems Approach},
year = {2013},
issue_date = {Spring 2013},
publisher = {John Wiley \& Sons, Inc.},
address = {USA},
volume = {34},
number = {1},
issn = {0738-4602},
url = {https://doi.org/10.1609/aimag.v34i1.2449},
doi = {10.1609/aimag.v34i1.2449},
journal = {AI Mag.},
month = mar,
pages = {67–77},
numpages = {11}
}

@inproceedings{cavazza2002interacting,
  title={Interacting with virtual characters in interactive storytelling},
  author={Cavazza, Marc and Charles, Fred and Mead, Steven J},
  booktitle={Proceedings of the first international joint conference on Autonomous agents and multiagent systems: part 1},
  pages={318--325},
  year={2002}
}

@article{mateas2002behavior,
  title={A behavior language for story-based believable agents},
  author={Mateas, Michael and Stern, Andrew},
  journal={IEEE Intelligent Systems},
  volume={17},
  number={4},
  pages={39--47},
  year={2002},
  publisher={IEEE}
}

@InProceedings{ryan2008interactive,
author="Ryan, Marie-Laure",
editor="Spierling, Ulrike
and Szilas, Nicolas",
title="Interactive Narrative, Plot Types, and Interpersonal Relations",
booktitle="Interactive Storytelling",
year="2008",
publisher="Springer Berlin Heidelberg",
address="Berlin, Heidelberg",
pages="6--13",
isbn="978-3-540-89454-4"
}

@inproceedings{yang2025steering,
  title={Steering Narrative Agents Through a Dynamic Cognitive Framework for Guided Emergent Storytelling},
  author={Yang, Chen and Gross, Markus and Wampfler, Rafael},
  booktitle={Proceedings of the AAAI Conference on Artificial Intelligence and Interactive Digital Entertainment},
  volume={21},
  number={1},
  pages={377--387},
  year={2025},
  doi={10.1609/aiide.v21i1.36841},
  url={https://doi.org/10.1609/aiide.v21i1.36841}
}

@misc{suckup2025,
  author = {{Proxima}},
  title = {Suck Up!},
  howpublished = {Video game},
  publisher = {Proxima},
  year = {2025},
  url={https://www.playsuckup.com/},
  note = {Played on PC}
}

@misc{portopia2023,
  author = {{Square Enix}},
  title = {SQUARE ENIX AI Tech Preview: THE PORTOPIA SERIAL MURDER CASE},
  howpublished = {Video game},
  publisher = {Square Enix},
  year = {2023},
  note = {Played on PC}
}

@misc{herstory2015,
  author = {Barlow, Sam},
  title = {Her Story},
  howpublished = {Video game},
  publisher = {Sam Barlow},
  year = {2015},
  note = {Played on PC}
}

\end{document}